\documentclass[letterpaper, 10 pt, conference]{ieeeconf}  

\IEEEoverridecommandlockouts                              
\usepackage{graphicx}		
\usepackage{booktabs}       
\usepackage[detect-all,per-mode=symbol]{siunitx}
\usepackage{amsmath}	
\usepackage{amssymb}	
\usepackage{commath}                    
\usepackage[capitalize]{cleveref}		
\usepackage{multirow}

\newcommand{\xtt}{x_{t+\Delta t}}
\newcommand{\ztt}{z_{t+\Delta t}}
\newcommand{\xtn}{x_t^{\mathrm{n}}}
\newcommand{\xte}{x_t^{\mathrm{e}}}
\newcommand{\xn}{x^{\mathrm{n}}}
\newcommand{\xe}{x^{\mathrm{e}}}
\newcommand{\xth}{x_t^{\mathrm{h}}}
\newcommand{\thetae}{\theta^{\mathrm{e}}}
\newcommand{\thetan}{\theta^{\mathrm{n}}}
\newcommand{\se}{\textrm{SE}(2)}
\newcommand{\Xd}{X_d}
\newcommand{\Zd}{Z_d}
\newcommand{\Thetad}{\Theta_d}

\newcommand{\kk}{k}
\newcommand{\seq}{\,{=}\,} 

\newcommand{\phin}{\phi^\textrm{n}}
\newcommand{\phinr}{\widetilde{\phi}^\textrm{n}}
\newcommand{\phir}{\widetilde{\phi}}

\newcommand{\kl}{\|}

\title{\LARGE \bf
Using Neural Networks to Generate \\ Information Maps for Mobile Sensors
}

\author{Louis Dressel and Mykel J. Kochenderfer
\thanks{This work was supported NSF grant DGE-114747.}
\thanks{The authors are with the Department of Aeronautics and Astronautics at Stanford University, Stanford, CA, 94305 USA. Email:
		{\tt\small \{dressel,mykel\}@stanford.edu}}%
}

\begin{document}

\maketitle
\thispagestyle{empty}
\pagestyle{empty}

\begin{abstract}

Target localization is a critical task for mobile sensors and has many applications.
However, generating informative trajectories for these sensors is a challenging research problem.
A common method uses information maps that estimate the value of taking measurements from any point in the sensor state space.
These information maps are used to generate trajectories; for example, a trajectory might be designed so its distribution of measurements matches the distribution of the information map.
Regardless of the trajectory generation method, generating information maps as new observations are made is critical.
However, it can be challenging to compute these maps in real-time.
We propose using convolutional neural networks to generate information maps from a target estimate and sensor model in real-time.
Simulations show that maps are accurately rendered while offering orders of magnitude reduction in computation time.

\end{abstract}

\section{INTRODUCTION}

Mobile sensing tasks are critical robotic applications in which a sensing agent gathers information about an environment.
In localization, a type of mobile sensing task, the information gathered by the agent reduces uncertainty about unknown parameters.
These parameters might represent the location of a target that must be found.
Example targets include GPS jammers~\cite{Perkins2016}, avalanche beacons~\cite{hoffmann2006}, or radio-tagged wildlife~\cite{cliff2015}.
Localizing these targets quickly is often critical, so planning informative paths for mobile sensors is an important research topic.

Unfortunately, planning paths that maximize gathered information is a difficult controls problem.
Long-term optimality can theoretically be achieved with dynamic programming, but these approaches are often computationally infeasible, and it is difficult to attain approximately optimal solutions~\cite{icaps2017}.
Greedy optimizations that maximize the information gathered in the next timestep are computationally feasible and have been implemented in many tasks~\cite{hoffmann2006,cliff2015,gnc2018}.
However, greedy planners can lead to poor long-term results and are vulnerable to unmodeled noise~\cite{icaps2017,millerthesis2}.

One approach to the planning problem is to generate an information map.
This map associates a state in the sensing agent's state space to some measure of information.
The agent can plan a path through this distribution of information that avoids getting stuck in local minima.
The information map is constructed using the target estimate, the sensor model, and information-theoretic quantities like Fisher information or mutual information.

Information maps are an integral component of ergodic control for mobile sensing tasks.
In this control framework, mobile sensors execute trajectories that are ergodic with respect to an information map.
Trajectories are ergodic if they spend time in a state space region proportional to the information available there.
There is empirical evidence that ergodic trajectories efficiently gather information while being robust to unmodeled sensor noise~\cite{millerthesis2}.

Unfortunately, these information maps can themselves be computationally challenging to generate.
As observations are made, the belief---the distribution over target locations---changes, and the information map changes as a result.
In order to incorporate these changes, new trajectories are planned and executed in a model-predictive fashion.
Before these trajectories can be re-computed, the information map must be updated with recent measurements; it is therefore crucial that these maps be generated in real-time.
However, the information map can be computationally expensive to update, hindering real-time application.

We propose using convolutional neural networks to generate information maps directly from beliefs.
These networks are trained offline using simulated trajectories and a sensor model.
When given a belief, these networks output the information maps or their Fourier coefficients.
As a result, the information maps are produced quickly and can easily be applied online in real mobile sensing tasks.
We demonstrate the speed improvements in simulations.
Depending on the sensor model and type of information map, computation time is reduced by two orders of magnitude.


\section{BACKGROUND}
\label{sec:background}

\subsection{Neural Networks}
The idea of using machine learning to speed up online computation is not new.
In an early example, support vector machines were used to determine if a robotic agent could reach other points in the state space~\cite{allen2014}.
Traditional approaches numerically solved a two-point boundary value problem to determine reachability, but this machine learning approach drastically reduced computation time.

In our application, we also have values with high computational complexity that must be computed in real-time.
We also try a machine learning approach but use convolutional neural networks instead of support vector machines.
Convolutional neural networks are a natural choice because our input is an information distribution over the state space.
This input is like an image and we expect there to be some spatial correlation between points in the state space.

Convolutional neural networks have had stunning success classifying and modifying images, in large part because of spatial correlation in images~\cite{krizhevsky2012,}.
A convolutional layer is a set of filters that is convolved over the input image.
These filters detect repeated features in the input.
Typically, a few convolutional layers are stacked until fed into a fully connected layer for the output.



\subsection{Ergodic Control}

One application that has seen extensive use of information maps is ergodic control for mobile sensors.
The mobile sensor maintains an information map $\phi$, which shows how information is distributed over its state space.
The sensor then plans a trajectory that is ergodic with respect to this distribution.
As the sensor executes the trajectory and collects measurements, the information map is updated and new trajectories are computed in a model-predictive fashion.

The sensor trajectory is converted into a spatial distribution $c$ over the state space $X$.
If $q$ is a sensor trajectory of duration $T$, the density of this spatial distribution at a point $x\in X$ is
\begin{equation}
c(x) = \int_0^T \delta(x - q(t)) dt\text,
\end{equation}
where $\delta$ is the Dirac delta function.

In some methods, an ergodic trajectory is generated by comparing $c$ to $\phi$ using a metric like KL divergence~\cite{ayvali2017}.
However, it is more common to decompose $c$ and $\phi$ into Fourier coefficients and modify the trajectory until the coefficients are roughly equal~\cite{mathew2011,miller2013,mavrommati2018}.
The information map $\phi$ is decomposed according to
\begin{equation}
\phi_\kk = \int_X \phi(x) F_\kk(x) dx\text,
\end{equation}
where $F_\kk$ is a Fourier basis function and $\kk$ is a multi-index with as many dimensions as the state space.
For example, if $X\subset \mathbb{R}^2$, $\kk = [k_1,k_2]$.
The highest-order coefficient in any dimension is $K$; in the $\mathbb{R}^2$ example, $k_1$ and $k_2$ each range from 0 to $K$.
In Euclidean space, the basis function $F_\kk$ is cosine-based~\cite{mathew2011}, but in special Euclidean groups such as $\se$ a basis function that uses the Bessel function and complex exponentials is used~\cite{miller2013b}.

The coefficients $c_\kk$ are derived in a similar fashion and the ergodic objective $\mathcal{E}$ compares the coefficients:
\begin{equation}
    \mathcal{E}(x) = \sum_\kk \Lambda_\kk \norm{c_\kk - \phi_\kk}_2^2\text.
\end{equation}
The weighting factor $\Lambda_\kk$ assigns higher weight to low frequency coefficients and is well explained in the literature~\cite{mathew2011}.
A trajectory is deemed ergodic when $\mathcal{E}$ is low.

Executing ergodic control in real-time can be challenging~\cite{mavrommati2018}.
One of the challenges is that the information map $\phi$ changes with new measurements.
An ergodic trajectory is generated for $\phi$, but this map becomes obsolete as new measurements are made.
Recomputing the information map for each new measurement are made can be computationally expensive and might not be feasible on a robot with limited computing power.
It can also be difficult to generate the Fourier coefficients $\phi_\kk$.
For this reason, we use neural networks not just to generate $\phi$ but also $\phi_\kk$.

\section{MODEL}
\label{sec:model}

As a motivating example, we consider the localization of a single, stationary target with a mobile sensor.

\subsection{Dynamic Model}

The stationary target has a location $\theta\in\Theta$.
The set of possible target locations $\Theta\subset\mathbb{R}^2$ is a 2D square.
The location $\theta$ consists of north and east components so that $\theta = [\thetan, \thetae]$.

At time $t$, the mobile sensor is in state $x_t\in X$.
The sensor state space depends on the sensor model used.
If the agent's heading is unimportant, then $X\subset\mathbb{R}^2$.
If heading matters, then the agent state space is a subset of a special Euclidean group: $X\subset\se$.
In this case, the agent state includes a heading in addition to its 2D position.

This paper's focus is on information map generation, so we use a simple dynamic model.
The mobile sensor has deterministic, single integrator dynamics.

\subsection{Sensor Models}
\label{sec:sensors}

Measurements are made every $\Delta t$ seconds.
At time $t$, the mobile sensor makes the measurement $z_t\in Z$, where $Z$ is the domain of possible measurements.

The sensor model provides $P(z_t \mid x_t, \theta)$, the probability of receiving measurement $z_t$ given mobile sensor state $x_t$ and target location $\theta$.
This probability is used in the filtering and estimation as well as generation of the information map.

We consider two sensor models in this work.
The first is a bearing-only sensor that returns bearing estimates to the target.
Such measurements can be obtained with beam-steering~\cite{Perkins2017}.
Because the beam is electronically steered, the sensing agent's heading does not affect the sensing model, resulting in the state space $X\subset \mathbb{R}^2$.
The sensor state consists of a north and east component: $x_t = [\xtn, \xte]$.
The measurement obtained at time $t$ is
\begin{equation}
    z_t = \beta_t + w_t\text,
\end{equation}
where $w_t$ is zero-mean Gaussian noise.
Measured east of north, the bearing $\beta_t$ to the target is
\begin{equation}
   \beta_t = \arctan\left(\frac{\thetae - \xte}{\thetan - \xtn}\right)\text.
   \label{eq:bearing}
\end{equation}
The probability $P(z_t \mid x_t, \theta)$ is derived from this model.

The second sensor model is a field-of-view (FOV) sensor introduced in previous work~\cite{gnc2018}.
The sensing agent makes radio strength measurements simultaneously with two directional antennas.
One points forward and the other rearward.
Only two measurements are possible so that $Z = \{0,1\}$.
A measurement of 1 is received when the front antenna measures a higher strength than the rear antenna; otherwise 0 is received.
Because the antennas are directional, we expect a measurement of 1 when the mobile sensor points at the target.
Here, the sensor's heading $\xth$ affects the measurement, so $X\subset\se$.
Denoting the sensor state $x_t = [\xtn, \xte, \xth]$, the measurement function is
\begin{equation}
	P(z_t = 1\mid x_t, \theta) = \begin{cases}
        0.9, & \text{if }\beta_t - \xth \in \left[-\ang{60},\ang{60}\right]\\
        \hfil 0.1, & \text{if }\beta_t - \xth \in \left[\ang{120},\ang{240}\right] \\
		\hfil 0.5, & \text{otherwise.}
	\end{cases}
	\label{eq:sensor}
\end{equation}

\subsection{Belief and Filtering}

Because the target location $\theta$ is unknown, a distribution over possible target locations is maintained.
This distribution is called the belief and $b_t$ denotes the belief at time $t$.


In this work, we use a discrete filter, sometimes called a histogram filter~\cite{thrun2005}.
In a discrete filter, the search area is discretized into grid.
The belief is an array representation of this grid.
The weight of an entry is the probability the target is in the corresponding grid cell.

Unlike Kalman filters, discrete filters do not require unimodal, Gaussian beliefs or linearizable sensing models.
Unlike particle filters, discrete filters represent the belief as an array, making it easier to feed into machine learning techniques like neural networks.
The discrete filter has also been used extensively in localization tasks~\cite{cliff2015,gnc2018}.

The term $b_t(\theta_i)$ is the probability the target is in cell $\theta_i$, according to the belief at time $t$.
Given the belief from the previous timestep, $b_{t-\Delta t}$, and the measurement made at the current timestep, $z_t$, the belief is updated according to
\begin{equation}
    b_{t}(\theta_i) \propto b_{t-\Delta t}(\theta_i) P(z_t \mid x_t, \theta_i)\text.
\end{equation}
The belief is normalized so it sums to one.

\section{INFORMATION MAPS}
\label{sec:maps}

The information map $\phi:X\to\mathbb{R}$ maps the state space to a measure of information quality or quantity.
To feasibly compute the map, we compute the information values at a discrete set of points $\Xd\subset X$.
The information at $x\in\Xd$ is computed using quantities like mutual or Fisher information.

Computing information values typically requires integrating over the target space $\Theta$, so we also limit this to a discrete set of points $\Thetad$.
It is sometimes necessary to integrate over the measurement space $Z$.
If this space is continuous, we also limit ourselves to a discrete set $\Zd$.
In bearing-only localization, where $Z=[\ang{0},\ang{360})$, we choose $\Zd = \{\ang{0},\ang{10},\dots,\ang{350}\}$.

\subsection{Mutual Information}
Mutual information is often used to guide mobile sensors in localization tasks~\cite{hoffmann2006,gnc2018}.
The mutual information at a state is equal to the expected reduction in belief entropy resulting from a measurement there.
Entropy captures the uncertainty in a distribution or random variable.
Because the goal in localization is to reduce uncertainty about the unknown parameter $\theta$, minimizing belief entropy is sensible.

Given two random variables $A$ and $B$, the mutual information $I(A; B)$ is the amount of information obtained about one variable given the other is known.
Equivalently, $I(A;B)$ gives the reduction in uncertainty of $A$ given $B$ is known or the reduction in uncertainty of $B$ given $A$ is known.
Mutual information is symmetric so $I(A;B) = I(B;A)$.

In localization, we are often interested in $I(b_t; \ztt)$, the reduction in uncertainty of $b_t$ given the next measurement is known.
Strictly speaking, $b_t$ is a distribution and not a random variable; we abuse notation and use $b_t$ to refer to the random variable describing the value of $\theta$, which has distribution $b_t$.
The measurement at the next timestep, $\ztt$, is a random variable because it is an unknown quantity.

The value of $I(b_t; \ztt)$ is made explicit in the relation
\begin{equation}
I(b_t; \ztt) = H(b_t) - H(b_t \mid \ztt)\text.
\end{equation}
The quantity $H(b_t)$ is the current entropy of $\theta$.
The quantity $H(b_t \mid \ztt)$ is the conditional entropy of $\theta$ given the next measurement were known.
We leverage the symmetry of mutual information:
\begin{equation}
    I(\ztt; b_t) = H(\ztt) - H(\ztt \mid b_t)\text.
    \label{eq:mi}
\end{equation}
In greedy control, \cref{eq:mi} is evaluated for each $\xtt$, or possible agent state at the next timestep.
The first term is
\begin{equation}
    H(\ztt) = -\sum_{z\in\Zd} P(\ztt\seq z)\log P(\ztt\seq z)\text,
\end{equation}
where, using total and conditional probability,
\begin{equation}
P(\ztt \seq z) = \sum_{\theta_i\in\Thetad}P(\ztt\seq z \mid \xtt, \theta_i)b_t(\theta_i).
\end{equation}
The second term in \cref{eq:mi} is
\begin{equation}
    H(\ztt \mid b_t) = \sum_{\theta_i\in\Thetad} b_t(\theta_i) \sum_{z\in\Zd}P_{zx\theta} \log P_{zx\theta}\text,
\end{equation}
where $P_{zx\theta} = P(\ztt\seq z \mid \xtt, \theta_i)$ for short.

When generating an information map, \cref{eq:mi} is evaluated for each $x\in \Xd$ instead of $X_{t+\Delta t}$, the list of states that can be reached in the next timestep.
Each term in \cref{eq:mi} is of order $O(\abs{\Thetad}\abs{\Zd})$ so generating the entire map is $O(\abs{\Xd}\abs{\Thetad}\abs{\Zd})$.
Each operation requires a call to the sensor model $P(z_t \mid x_t, \theta)$, which can be expensive.
For example, a bearing sensing modality requires calls to relatively expensive trigonometric functions.
However, these calls can be reduced with caching and memoization.

The main computational concern is that the numbers of sensor and target states are often exponential functions of some other variable.
Consider a mobile sensor localizing a target in a square field.
We might discretize to $n$ states per dimension, meaning the sensor and target could each occupy any of $n^2$ states.
Generating the information map is of order $O(n^4\abs{Z})$.
Clearly, increasing the discretization or the size of the search area incurs enormous increases in computation.

\subsection{Fisher Information}
Fisher information offers another way to generate information maps.
The Fisher information $\mathcal{I}(\alpha)$ describes the amount of information that a random variable carries about the unknown parameter $\alpha$.

In our case, the observable variable is the measurement $z$ and it is conditional on the sensor state $x$ and target state $\theta$.
We restrict ourselves to the bearing-only sensor model, where the measurement value is scalar and has Gaussian noise that is constant across the state space.
The Fisher information matrix for a specific sensor-target state is
\begin{equation}
    \mathcal{I}(x, \theta) = \frac{1}{\sigma^2}\nabla_{\theta} g(\theta, x)\nabla_{\theta}g(\theta, x)^\top\text,
    \label{eq:fim}
\end{equation}
where the $\sigma$ is the standard deviation of the Gaussian noise and $\nabla_{\theta}g(\theta,x)$ is the gradient of the measurement function $g$ with respect to $\theta$.
In bearing-only sensing, $g$ is the true bearing in \cref{eq:bearing} and its gradient is
\begin{equation}
    \nabla_{\theta}g(\theta,x) = \frac{1}{\norm{\theta - x}^2}
    \begin{bmatrix}
        \thetae - \xe\\
        \xn - \thetan
    \end{bmatrix}\text.
    \label{eq:gradient}
\end{equation}
When calculating Fisher information for a point in the sensor's state space, the sensor state is known but the target state is not.
Therefore, the current belief is used to take an expectation over sensor states:
\begin{equation}
    \Phi(x) = \sum_{\theta_i\in\Thetad} b_t(\theta_i)\mathcal{I}(x,\theta_i)\text.
    \label{eq:Phi}
\end{equation}
An information map requires a scalar value of information at each point, so the determinant is commonly used~\cite{miller2016}:
\begin{equation}
    \phi(x) = \det \Phi(x)\text.
    \label{eq:det}
\end{equation}
The information map $\phi$ is built using \cref{eq:fim,eq:gradient,eq:Phi,eq:det} to evaluate the information at each point $x\in\Xd$.

The computational complexity of generating the Fisher information map is $O(\abs{\Xd}\abs{\Thetad})$, better than mutual information by the factor $\abs{\Zd}$.
This factor is eliminated in part because of the simplified version of Fisher information in \cref{eq:fim}.
In cases with more complex noise models, integration over the measurement space is needed to compute $\mathcal{I}(\theta, x)$.
However, $\mathcal{I}(\theta, x)$ can be precomputed offline, so that the Fisher information map complexity is still $O(\abs{\Xd}\abs{\Thetad})$.
Further, there are no calls to the log or measurement functions.
The low complexity helps explain why Fisher information maps are common in prior work~\cite{millerthesis2,miller2016}, including a real-time implementation on a robot~\cite{mavrommati2018}.

However, Fisher information is not the best metric for all problems, and picking the exact form of the information map is an open research question.
In some problems, it is not even clear how to apply Fisher information.
For example, the FOV sensor model from \cref{sec:sensors} has just two discrete observations, and the gradient is not well defined.
Mutual information might be more appropriate in that case.

\section{NETWORK DESIGN}
\label{sec:network}

We design networks for a mobile sensor according to the models from \cref{sec:model}.
A stationary target sits in a $\SI{200}{\meter} \times \SI{200}{\meter}$ field.
The belief is represented with an $n\times n$ discrete grid, where $n=28$.
The weight of each cell in the belief gives the probability the target is in the grid.
The belief is initialized to a uniform distribution.

The agent moves through this field while searching for the target.
When using the bearing modality, we also discretize the agent state space to $n\times n$ points in the search area.
When using the FOV modality, the agent state space is $n\times n\times 36$, as we discretize possible agent headings into 36 points.

By using the sensor models and mutual or Fisher information, information maps over the agent state space can be generated.
In the bearing modality, these maps cover $n\times n$ points; in the FOV modality, they cover $n\times n\times 36$ points.

We also generate Fourier coefficients from these information maps.
For the bearing modality, we use $K=5$ as the highest order coefficient, in line with prior work~\cite{mavrommati2018}.
In the FOV modality, we use $K=17$, the smallest value that captured major features in observed information maps.

\subsection{Neural Network Architectures}


The first architecture takes in an $n\times n$ discrete belief and outputs either an $n\times n$ or an $n\times n\times 36$ information map, depending on the sensing modality.

The input passes through two convolutional layers, a fully connected layer, a deconvolution layer, and a softmax activation.
The number of filters per convolutional layer depends on the size of the output.
The softmax activation ensures the output sums to one, making it easier to use KL divergence as the loss function.

The second architecture takes in an $n\times n$ belief and outputs a vector containing the coefficients of the information map.
Because there are far fewer coefficients than points in the belief, the network is simpler.
The network consists of two convolutional layers before two fully connected layers.
The mean absolute error is used as the loss function.

\subsection{Training}

To train the networks, we run 500 simulations of 20 steps each.
In each simulation, the target sits at a random location.
The sensing agent selects its control input with a one-step, mutual information optimization.
Measurements are made at each step, after which an information map is generated and decomposed into Fourier coefficients.
The beliefs are used as training inputs, and the resulting maps and coefficients are used as training outputs.

Training was done on a Tesla k40c graphics processing unit (GPU).
A GPU is not necessary, but it reduced training time from a few hours to about ten minutes.

Overfitting is always a concern with machine learning.
To minimize overfitting, we separate 10\% of the training data into a validation set.
At each epoch, the loss is evaluated on both the training and validation sets.
If loss diverges, overfitting has likely occurred.
This behavior was not observed.

\subsection{Complexity in Evaluation}
\label{sec:complexity}

Before evaluating the trained networks in simulation, we consider the computational complexity of these evaluations.
The networks are trained offline, so it does not matter if training is slow.
However, a trained network must generate information maps from beliefs in real-time.
The computational complexity of evaluating a convolutional neural network for a new input is
$O( \sum_{l=1}^d n_{l-1} \cdot s_l^2 \cdot n_l \cdot m_l^2 )$,
where $d$ is the number of convolutional layers, $n_{l-1}$ is the number of input layers to layer $l$, $s_l$ is the filter width of layer $l$, $n_l$ is the number of filters in layer $l$, and $m_l$ is the width of layer $l$'s output~\cite{he2015}.
The input is 2D so the number of input layers is $n_0=1$.
We set the stride to one and zero-pad so that the output width $m_l$ equals the input width.
The input is an $n\times n$ belief, so the output width of a layer is $n$.

Because convolutions take most of the computation time, this complexity does not include the cost of any pooling or fully connected layers; prior empirical work suggests these layers account for 5--10\% computation time~\cite{he2015}.

If the network structure is held constant except for the input size $m_l = n$, then the asymptotic complexity is $O(m_l^2) = O(n^2)$.
Recall that Fisher information was $O(\abs{\Thetad}\abs{\Xd})$; if we use $n^2$ points for $\Thetad$ and $n^2$ for $\Xd\subset \mathbb{R}^2$, the asymptotic complexity is $O(n^4)$.
If $\Xd\subset \se$ is discretized with $n^3$ points, then the complexity is $O(n^5)$.
In theory, neural networks can generate information maps faster than computing them with Fisher or mutual information.

Of course, this result is theoretical and describes the limit as $n$ grows.
In reality, other network elements affect computation time.
Further, convolutional layers often have nonlinear activation functions at their output, which can be expensive to compute.
Finally, it is possible the network structure must implicitly grow with input width $n$.
Perhaps more filters would be needed to capture fine-scale details that appear due to finer discretization of the state space.

\section{SIMULATIONS}
\label{sec:simulations}

\begin{figure}
    \centering
    \includegraphics[trim={.79in .25in 1.1in .3in},clip,width=1.6in]{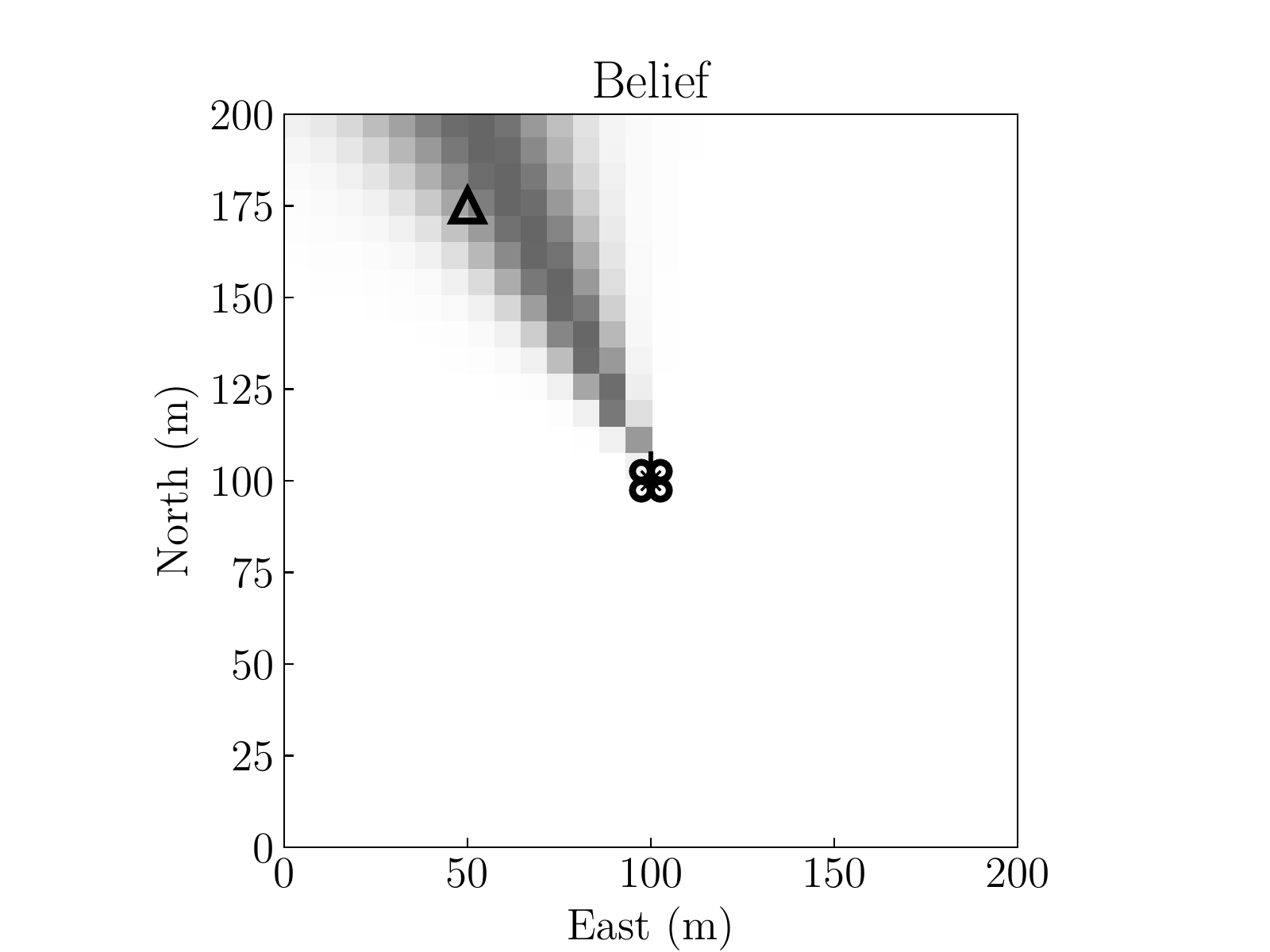}
	\includegraphics[trim={1.07in .25in 0.82in .30in},clip,width=1.6in]{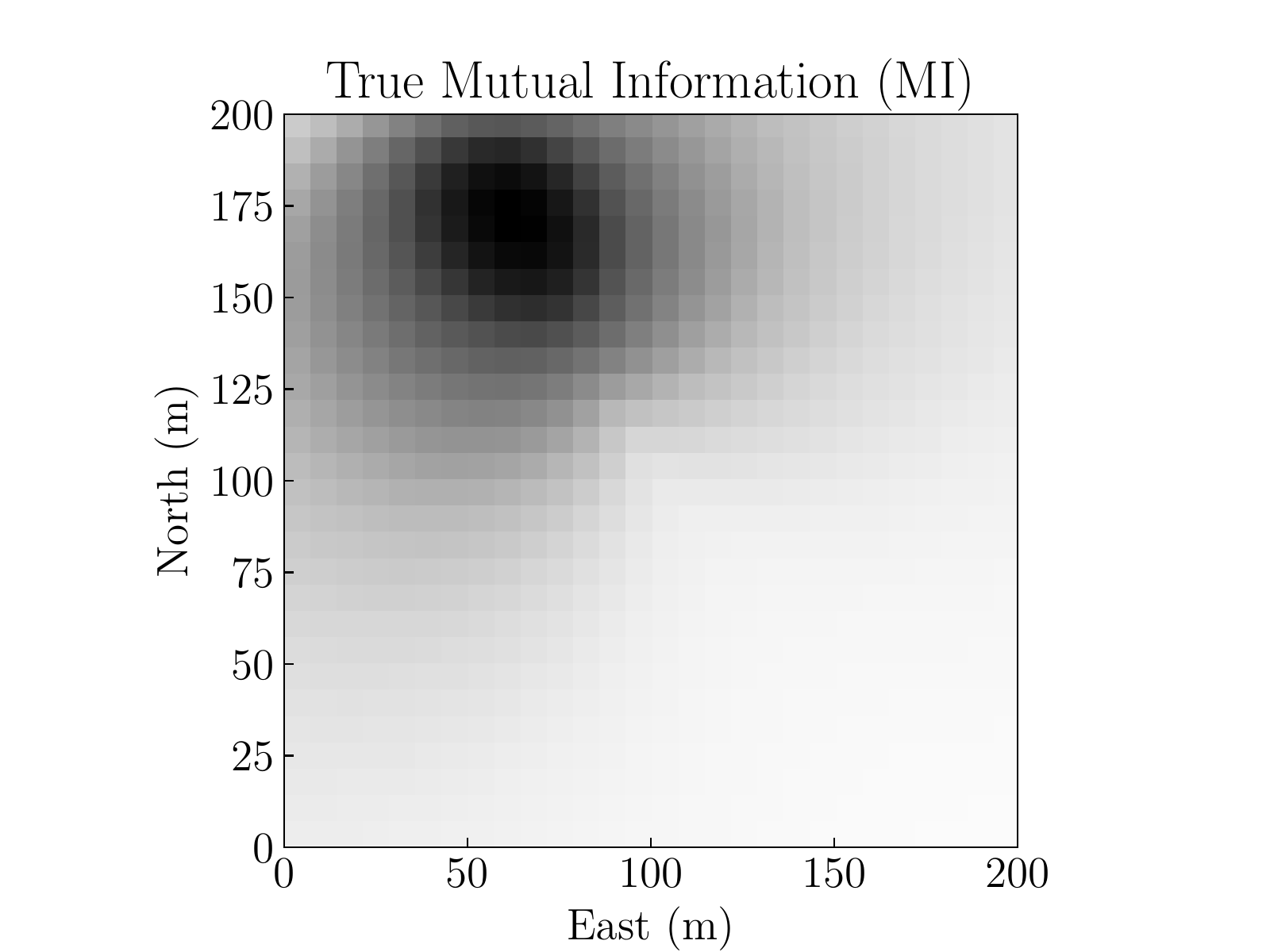}
	\includegraphics[trim={0.79in .0in 1.1in .20in},clip,width=1.6in]{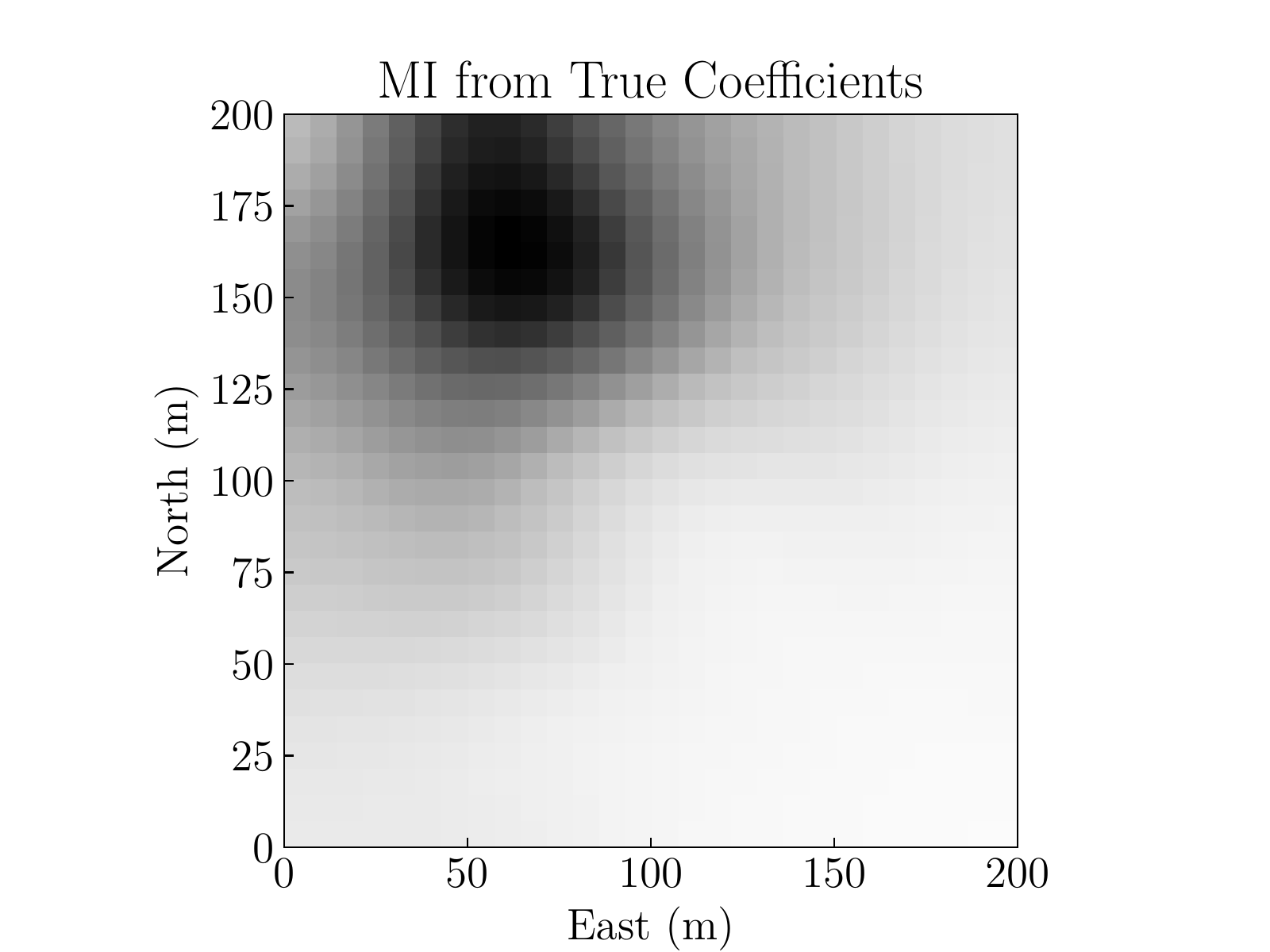}
	\includegraphics[trim={1.07in .0in 0.82in .20in},clip,width=1.6in]{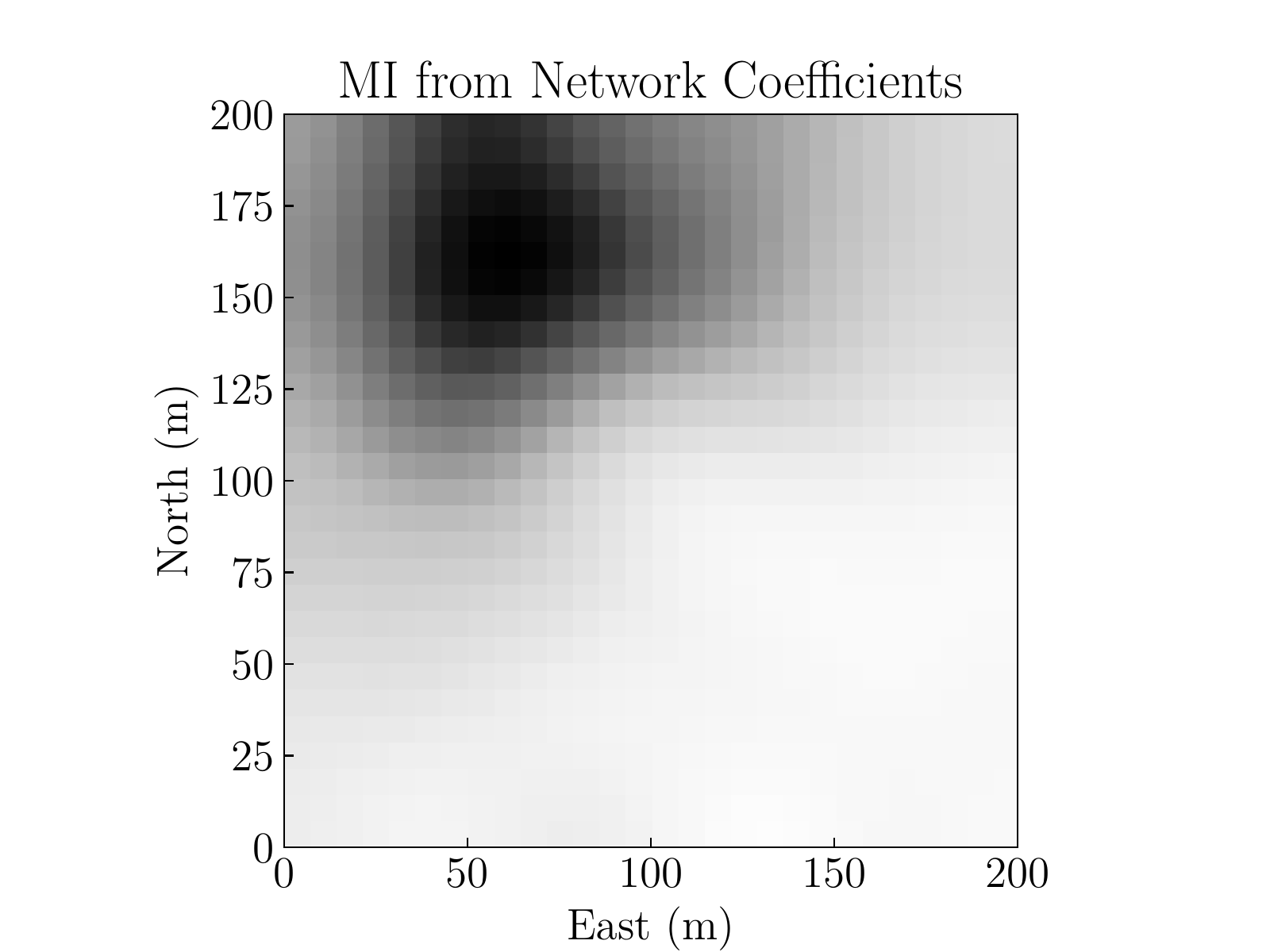}
    \caption{The mobile sensor (quadrotor) receives a bearing measurement to a target (triangle) and generates a belief. A mutual information map is then generated (upper right). A Fourier decomposition of this map is generated and the map is regenerated (bottom left). The Fourier coefficients generated by the neural network are also used to generate a map (bottom right).}
    \label{fig:bo_example}
\end{figure}

Once designed and trained, the networks are evaluated in simulations.
After each observation, the belief is updated and information maps are generated along with their Fourier coefficients.
These are compared to the neural network outputs.
An example is shown in \cref{fig:bo_example}.

All quantitative results in this section are from 100 20-step simulations with random target locations.
As in the data generation, the agent moves according to a myopic entropy minimization.
As a result, the beliefs seen in execution are similar to, but not necessarily equal to, those seen in training.

A tilde indicates a distribution was generated from Fourier coefficients, and the superscript $\text{n}$ indicates the distribution was generated by a neural network.
For example, $\phi$ is the true information map generated by the equations in \cref{sec:maps}; $\phir$ is the distribution generated from the true Fourier coefficients---that is, coefficients generated from the true distribution.
The distribution $\phinr$ is generated from the network-produced coefficients and $\phin$ is the neural network approximation of the true information map.

\subsection{Quality of Approximation}

Because neural networks are nonlinear function \textit{approximators}, there will be some degradation in the information maps produced.
KL divergence is used to evaluate this degradation quantitatively.
The KL divergence $D(P \kl Q)$ is a measure of how well $Q$ approximates $P$; the KL divergence is zero when $Q$ equals $P$.

\Cref{tab:kl} shows the average KL divergence after each simulation step.
The first quantity, $D(\phi\kl\phin)$, compares the network-produced information maps to the true maps.
The second quantity, $D(\phir\kl\phinr)$, captures the quality of the network-produced coefficients by comparing their reconstructed information map to that reconstructed from the true coefficients.
The third quantity, $D(\phi\kl\phir)$, compares the map generated from the true coefficients to the true information map.
Fourier coefficients introduce band-limiting degradation but are still used to guide mobile sensors~\cite{millerthesis2,mathew2011,miller2013,miller2013b,mavrommati2018}, so this last value is a useful reference of acceptable quality.

\begin{table}
    \centering
    \caption{Measuring Network Map Quality with KL Divergence.}
    \begin{tabular}{@{} llccc @{}}
        \toprule
        Modality & Metric & $D(\phi \kl \phin)$ & $D(\phir \kl \phinr)$ & $D(\phi \kl \phir)$\\
        \midrule
        \multirow{2}{*}{Bearing} & Fisher & 0.069 & 0.00045 & 2.78\\
        & Mutual & 0.036 & 0.0074 & 0.049\\
        \cmidrule{1-5}
        FOV & Mutual & 0.038 & 0.010 & 0.10\\
        \bottomrule
    \end{tabular}
    \label{tab:kl}
\end{table}

\begin{figure}
    \centering
    \includegraphics[trim={.79in .25in 1.1in .2in},clip,width=1.6in]{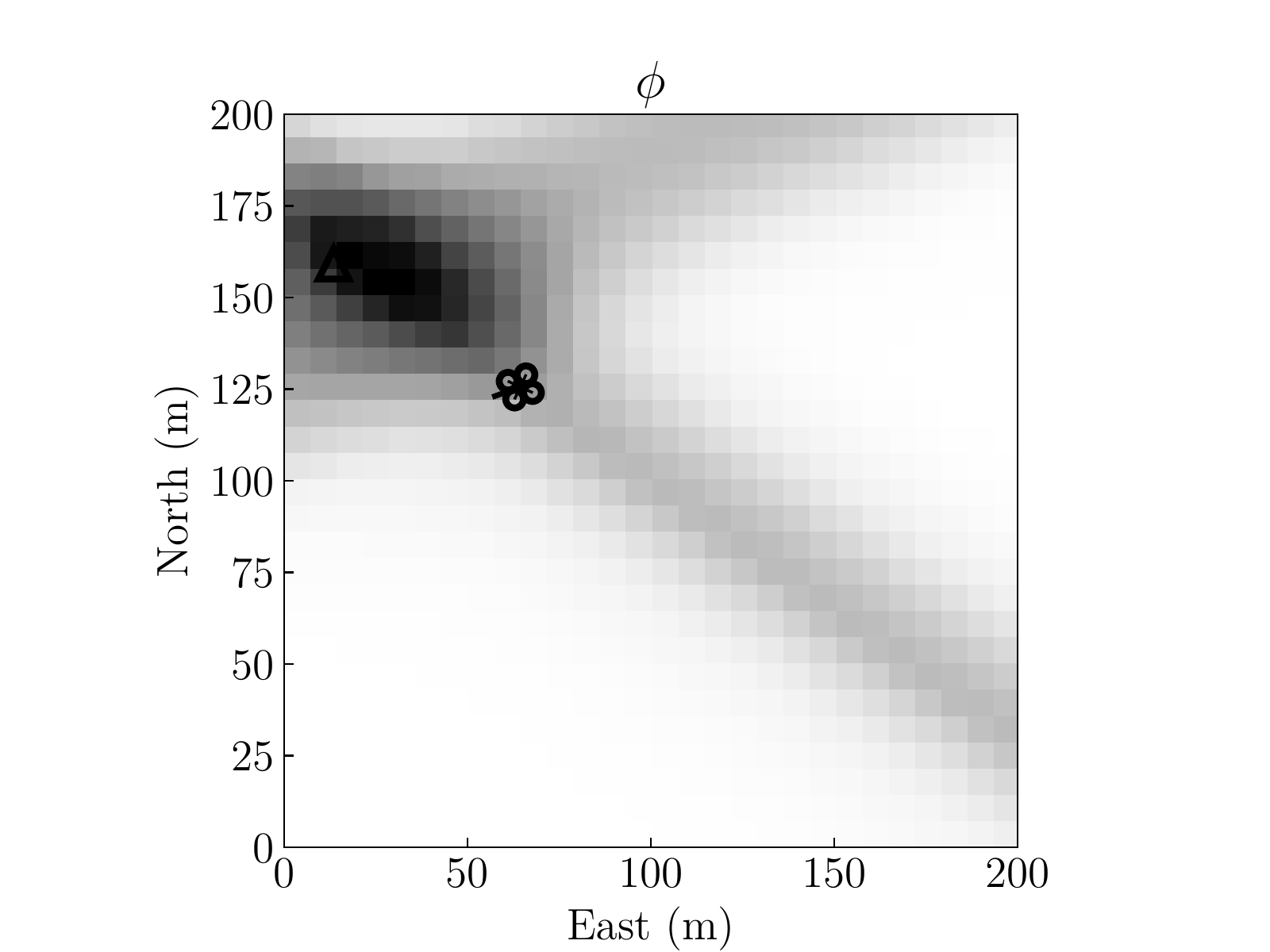}
	\includegraphics[trim={1.07in .25in 0.82in .20in},clip,width=1.6in]{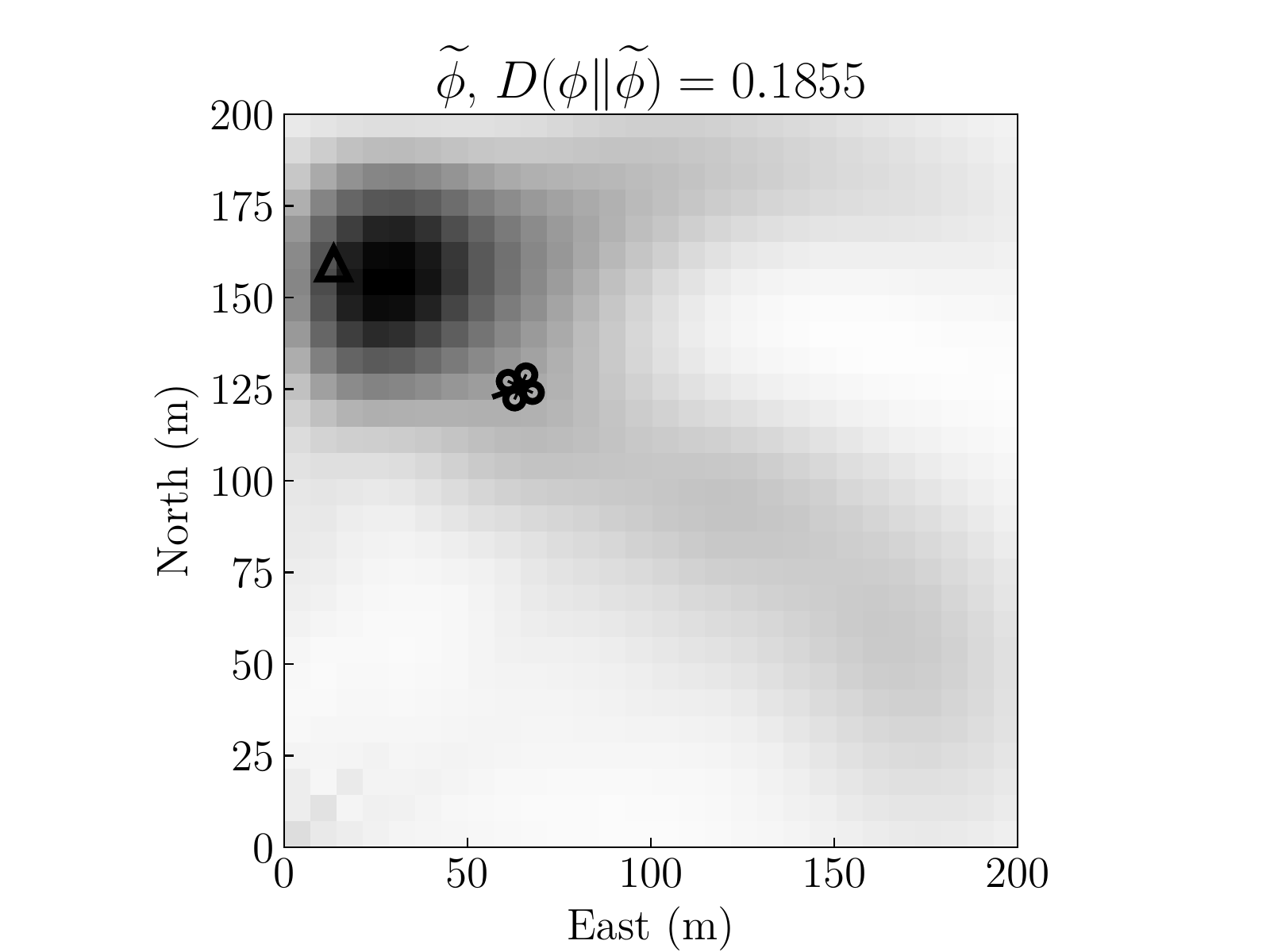}
	\includegraphics[trim={0.79in .0in 1.1in .20in},clip,width=1.6in]{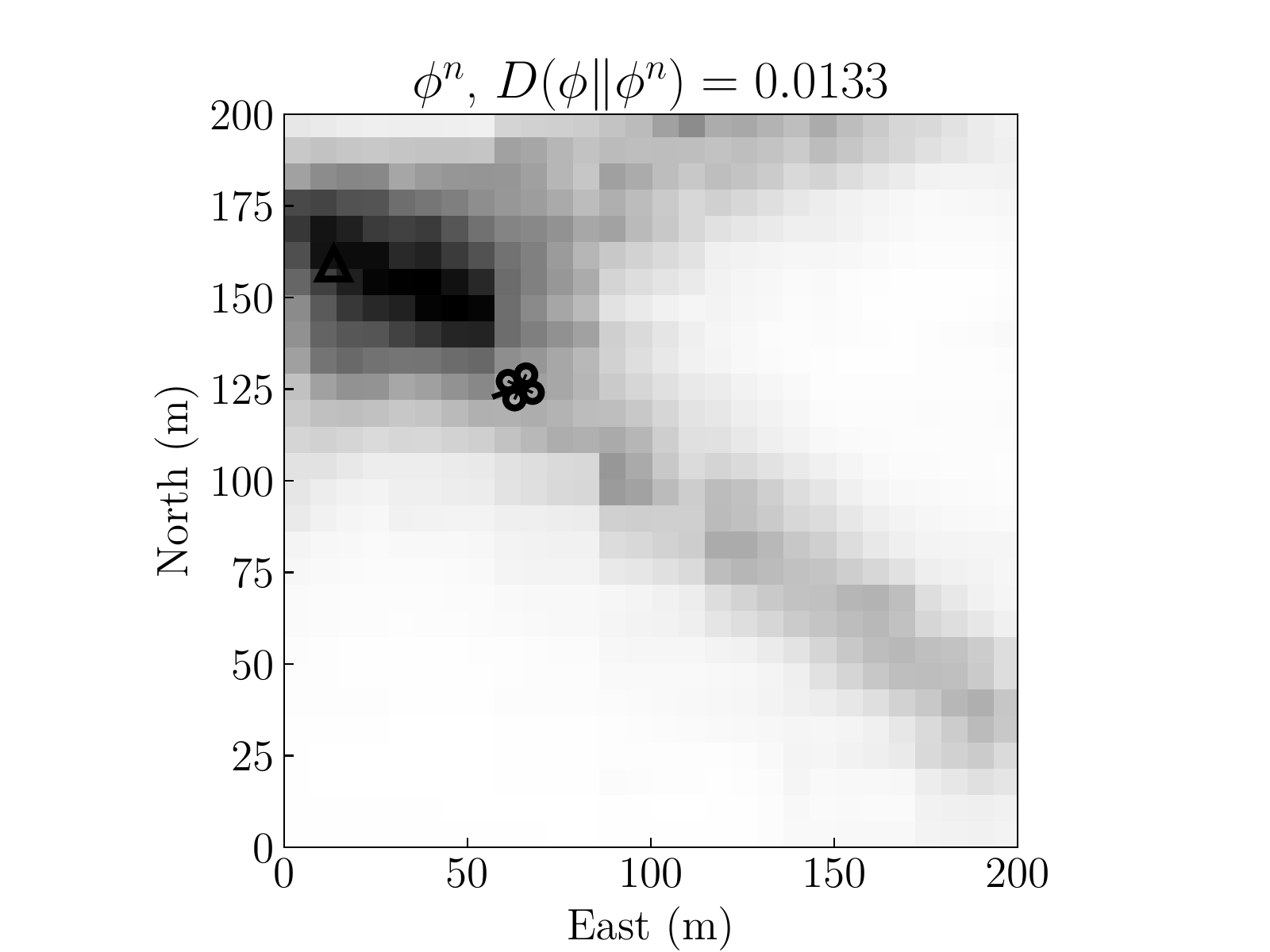}
	\includegraphics[trim={1.07in .0in 0.82in .20in},clip,width=1.6in]{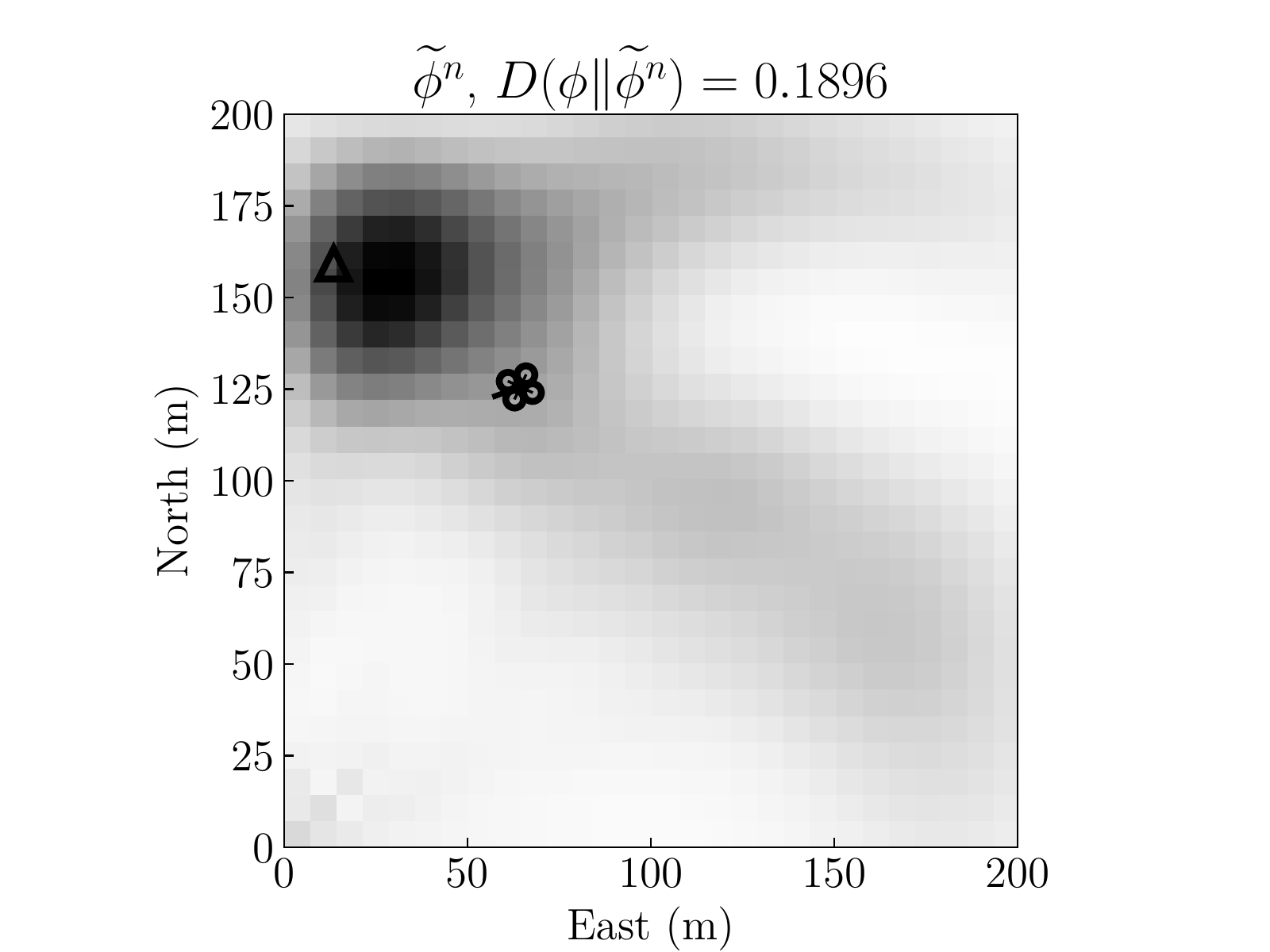}
    \caption{Comparison of true mutual information map and approximations during one timestep of FOV simulation. The information map covers $\se$, but a 2D slice at \ang{0} heading is shown here.}
    \label{fig:fov}
\end{figure}

The results suggest the networks accurately capture the information maps.
The divergence values between the true FOV maps and the network maps are low.
The divergence is only 0.038 when comparing the network map to the true map.
In comparison, the divergence is nearly triple that when using the true coefficients to reconstruct the information map, suggesting that more information is lost when approximating with the true coefficients than with the neural network.
If the true coefficients can be used in control tasks, then the network output will suffice as well.
\Cref{fig:fov} shows the approximations are also visually similar to the true maps.


\subsection{Computation Time}

\Cref{tab:time_k} shows the mean time to generate maps and coefficients from beliefs.
For the true methods, the map is made before decomposing it into coefficients, so the true coefficient time includes the true map generation time.

\begin{table}
    \centering
    \caption{Computation Time for True and Neural Network (NN) Maps.}
    \begin{tabular}{@{} lllll @{}}
        \toprule
        & & & \multicolumn{2}{c}{Time to Compute (s)}\\
        \cmidrule{4-5}
        Modality & Metric & Method & Map & Coefficients\\
        \midrule
        \multirow{4}{*}{Bearing} & 
            \multirow{2}{*}{Fisher}
                & True & 0.0061 & 0.0061\\
                & & NN & 0.0031 & 0.0016\\
        \cmidrule{2-5}
                & \multirow{2}{*}{Mutual} & True & 0.33 & 0.33\\
                                          & & NN & 0.0021 & 0.0013\\
        \cmidrule{1-5}
        \multirow{2}{*}{FOV} & \multirow{2}{*}{Mutual} 
                             & True & 0.76 & 1.33\\
         & & NN & 0.0093 & 0.026\\
        \bottomrule
    \end{tabular}
    \label{tab:time_k}
\end{table}

In the bearing modality, where the information map is a distribution over $\mathbb{R}^2$, the time to compute Fourier coefficients from the map is trivial.
Both the domain and number of coefficients are small, leading to fast computation.

Fisher information is also computed rapidly, resulting in computation times that are slower than, but comparable to, the neural network times.
Although a neural network can be much faster in the asymptotic limit, there is not much difference at the map size used in this work.

However, when using mutual information, neural networks generate maps and coefficients roughly two orders of magnitude faster.
This difference holds in the FOV modality, where the information maps cover $\se$.
The Fourier decomposition is slow because more coefficients are needed to faithfully represent the distribution and there is another dimension to integrate over.
The neural network is much faster.

Simulations were performed on a laptop computer with an i7 processor and 8 GB RAM.
Neural network evaluations were performed on the CPU (instead of the GPU) for a fair comparison.
Care was also taken to reduce the computation time of mutual information and its Fourier coefficients.
Julia, a high-level language whose performance approaches C, was used.
Caching and memoization were used to eliminate calls to measurement functions or the complex functions used in $\se$ Fourier decomposition.
Vectors were ordered to match Julia's column-major ordering and prevent cache misses.
Nonetheless, the neural network generated maps much more quickly and could be comfortably used at rates greater than $\SI{20}{\hertz}$, allowing real-time use.



\section{CONCLUSION}
\label{sec:conclusion}

Convolutional neural networks can generate high-fidelity information maps in real-time, allowing mobile sensors to update maps as new observations are made.
This technique is already being implemented on a real robot~\cite{gnc2018}, which uses the models in this paper.
Future work will evaluate the robustness and fidelity of network-generated maps in the presence of unmodeled noise or trajectories significantly different from those seen in training.
Other approximation techniques will also be compared to the neural network approach.

%

\bibliographystyle{IEEEtran}
\bibliography{bib}

%
%
%
%

\end{document}